\documentclass{ieeeaccess}
\usepackage{cite}
\usepackage{amsmath,amssymb,amsfonts}
\usepackage{algorithmic}
\usepackage{graphicx}
\usepackage{dblfloatfix}
\usepackage{textcomp}
\usepackage{subcaption}
\usepackage{array}
\usepackage{booktabs}
\definecolor{tan}{rgb}{0.8235, 0.7059, 0.5490} 
 \def\BibTeX{{\rm B\kern-.05em{\sc i\kern-.025em b}\kern-.08em
   T\kern-.1667em\lower.7ex\hbox{E}\kern-.125emX}}
\begin{document}

\title{Developing a Natural Language Understanding Model to Characterize Cable News Bias}

\author{\uppercase{Seth P. Benson}\authorrefmark{1}, 
\uppercase{Iain J. Cruickshank}\authorrefmark{2}, 
}
\address[1]{Carnegie Mellon University (e-mail: spbenson@andrew.cmu.edu)}
\address[2]{United States Military Academy (e-mail: iain.cruickshank@westpoint.edu)}

\markboth
{Benson and Cruickshank: Developing a Natural Language Understanding Model to Characterize Cable News Bias}
{Benson and Cruickshank: Developing a Natural Language Understanding Model to Characterize Cable News Bias}

\corresp{Corresponding author: Seth P. Benson (e-mail: spbenson@andrew.cmu.edu).}

\begin{abstract}
Media bias has been extensively studied by both social and computational sciences. However, current work still has a large reliance on human input and subjective assessment to label biases. This is especially true for cable news research. To address these issues, we develop an unsupervised machine learning method to characterize the bias of cable news programs without any human input. This method relies on the analysis of what topics are mentioned through Named Entity Recognition and how those topics are discussed through Stance Analysis in order to cluster programs with similar biases together. Applying our method to 2020 cable news transcripts, we find that program clusters are consistent over time and roughly correspond to the cable news network of the program. This method reveals the potential for future tools to objectively assess media bias and characterize unfamiliar media environments. 
\end{abstract}


\titlepgskip=10pt

\maketitle

\section{Introduction}
\label{sec:introduction}

The increasing trend of political polarization in the United States has garnered significant attention in recent literature. This trend, prominent at both national and local government levels, is reflected in media consumption patterns that indicate partisan polarization in the public \cite{hollander2008tuning}. Despite the plethora of studies on the subject, there is a noticeable gap in the literature regarding data-driven, computational analysis of the language used in political media sources. This void is particularly conspicuous in the study of cable news, which has been linked to the intensification of polarization. Furthermore, although computational research extensively explores differing sentiments towards issues, the application of cross-subject bias models remains limited.

In this paper, we introduce an application of advanced Natural Language Understanding techniques to quantify the partisan bias in cable news. Bias is delineated by two principal aspects: \textit{what} a source selects to discuss, and \textit{how} they choose to portray it. By leveraging named entity recognition, we identify critical topic words within transcripts, followed by stance analysis to ascertain the positive or negative framing of these topics. This approach facilitates an understanding of the diverse stances cable news programs adopt towards a set of topics, and how these programs differ in their choice of topics. Utilizing this information, we generate bias clusters of programs and examine their evolution over time. Our findings predominantly reveal consistent bias clusters strongly associated with a program's network.

The methodology outlined in this paper provides a more adaptable approach to characterizing bias compared to previous studies in the field. Notably, our technique does not necessitate controlling for the topic discussed in the analyzed text, enabling its broad application to various political content. The primary contributions of our research include:

\begin{itemize}
\item We have developed a novel technique for characterizing bias in cable news. This technique combines named entity recognition and stance analysis, providing a comprehensive view of both the subjects covered by a program and the stance taken on those subjects.

\item We have examined the consistency of program biases throughout 2020, providing a temporal analysis of bias in cable news. This analysis reveals how bias clusters evolve over time and how they are associated with the network of the program.

\item We have demonstrated the superiority of stance detection over sentiment analysis in determining bias. Our findings show that stance detection, which considers both the subject and the perspective taken on the subject, provides a more nuanced and accurate measure of bias compared to sentiment analysis.
\end{itemize}

\section{Related Research}
\label{sec:introduction}

\subsection{Cable News and Media Bias}

Despite the rise of the internet as a means of news consumption, cable news remains a prominent source of news and political information for the American public. Since 2006, daytime cable news viewership has seen a consistent increase alongside the entrance of primetime cable news as a major pull of viewers \cite{nw_cable_nodate}. Cable news channels have also been shown to have an impact on the opinions and political behavior of voters.\cite{martin_bias_2017} Taken together, this information provides justification for the study of cable news, which continues to hold a great deal of attention and influence in modern mass media. If, as previous research indicates, cable news has a tangible effect on viewers' political behavior, then understanding the biases present in the medium is especially important.

Media, including cable news, is broadly perceived as having political biases. Through the use of "expert reviewers" and public surveys, firms have assembled political bias ratings for news sources. \cite{noauthor_allsides_2019} They have found that most media sources, especially cable news sources, have a political lean to their coverage. This conclusion matches the view of the public. Americans' trust in the media to ``get the facts straight" and ``deal fairly with all sides" has been steadily declining.\cite{groeling_media_2013, inc_americans_2021} Cable news particularly has a poor perception, with less than 16 percent of the public having a ``great deal" or ``quite a lot" of trust in television news. This popular perception of cable news bias further invites a more rigorous characterization of bias on TV. However, current assessments of cable news bias based on "expert reviewers" or public polling are limited by their subjective nature. Ultimately, these methods rely on human assessment, which impairs truly objective determinations.

\subsection{Social Science Literature and Media Bias}

Social science has taken several approaches to examine bias in media and cable news specifically. The first of these is through looking at media \textit{gatekeeping bias}, or bias in the process of determining what or whom to cover \cite{padgett_as_2019}. On cable news, one way this can be assessed is through the figures that are brought on to be guests on shows. Some researchers have approached cable news bias by assessing the individual ideology scores of cable news speakers. One approach to this is using publicly available political donation data to map the ideological ideal points of individuals \cite{bonica2016avenues}. When analyzing the individuals that appeared on cable news, researchers found that cable news stations have a high amount of partisan bias in their guests, especially in primetime news slots \cite{kim_measuring_2022}. Another way to map gatekeeping bias in cable news is by analyzing the appearance of congressmembers. For each congressmember, ideological ideal points can be calculated through spatial analysis of congressional voting data. \cite{poole2007party} Using these ideological ideal points, research has found that ideologically extreme members are over-represented on cable news.\cite{padgett_as_2019, harmon_meet_2007} The media sources discussed on cable news can also reflect bias. Mapping sources cited in cable news reveals that channels of different partisan leans have distinct networks of news sources with little overlap, and an analysis of the think tanks cited on different cable news networks has demonstrated a preference for think tanks associated with the channels' partisan leans. \cite{conway-silva_reliable_2020, groseclose_measure_2005}

Social science literature has also made attempts to analyze language in news media, although they are oftentimes computationally limited. A large amount of media bias analysis in social science has been qualitative in nature, which allows for maximum interpretation without restricting the analysis to specific methodological techniques \cite{schreier2012qualitative}. From 1990 to 2005, qualitative analysis made up nearly half of social science media frame studies \cite{matthes_whats_2009}. Codebooks are one of the most widely used forms of quantitative analysis. One way codebooks are utilized is in manual coding, in which blind participants follow explicit rules to evaluate whether media content expresses certain frames toward subjects for several clips or articles \cite{bailard2016corporate}. Computer-assisted coding has also been completed through automated counts of keywords and supervised learning through training statistical models with previously coded content \cite{hillard2008computer, grimmer2013text}. One other approach has been centering resonance analysis, a form of network-based text analysis that "characterizes large sets of texts by identifying the most important words that link other words in the network" \cite{papacharissi_news_2008}. But, understanding the significance of the words identified in this method still requires human interpretation. While many social science studies leverage computational assistance, most are still dependent on human interpretation of text and leverage techniques that are well-suited only for a limited scope of analysis.

\subsection{Natural Language Understanding and Media Bias}

Hamborg et al. identify three main types of bias in news production: fact selection, writing style, and presentation style \cite{hamborg_automated_2019}. Fact selection is strongly related to the gatekeeping bias in social science literature. Presentation style relates to the visual bias of news presentation, including the bias in picture selection identified in social science. Finally, there is the bias introduced by the actual words and writing style used, or the writing style bias. Writing style bias is the the word choice --- or lexical bias --- and framing biases in the text; a news piece can be biased by both the words, or phrases, it chooses to use as well as the context of keywords in the text \cite{hamborg_automated_2019, DAlonzo:2022}.

Framing involves the language associated with different terms or issues, which is meant to inspire a particular effect in the consumer of that language. Through the analysis of hubs in formal mental networks, co-occurances of words within texts can be used to describe the viewpoints of text authors \cite{stella_forma_2020}. These networks perform dependency parsing to identify the syntactic relationship between words alongside their emotional perception and have been extended to understanding how media sources frame topics and figures \cite{semeraro_emotional_2022}. Additionally, moral framing, determined by a supervised learning coding model, can describe the moral perspectives present in different sources’ coverage (ex: Injustice) and demonstrate differences in moral perspectives between liberal and conservative sources. \cite{shahid_detecting_2020}. Finally, a recent sub-class of framing is \textit{Information Bias}, which is the conveyance of side information about the main event in the text in order to frame that main event in a certain way for the reader \cite{Fan:2019, Guo:2022}. Taken together, the framing of information, through various means, is an important determinant in the bias of a news piece and the effect the piece is meant to have on a consumer of the news piece.

Sentiment and affective analysis refer to computational practices that examine how positive or negative statements are towards their subjects. One way this can be done is through coded analysis of word affect scores. By coding the positive or negative affect of over 2,258 words and applying the developed lexicon to words surrounding key names, researchers have been able to determine the press favorability of political figures or candidates \cite{grefenstette_coupling_2004}. Recently, more advanced techniques have been explored. When controlling for topic by selecting articles discussing the same issue, Natural Language Tool Kit's VADER Sentiment Intensity Analyzer has been used to infer media source bias \cite{acharya_sentiment_2021}. The tool can identify the sentiment and political tone of different articles, which allows researchers to determine reporting differences between news organizations \cite{acharya_sentiment_2021}. 

Building off of this concept of press favorability, one combination of sentiment with framing was done by analyzing the affective scores of political news articles in relation to the partisan lean of the political figures they mention \cite{hamborg_newsalyze_2021}. This model allowed for bias identification that improved user bias awareness in an experimental setting. But, the increased number of speakers and issues discussed on cable news makes this model less equipped to analyze them. Additionally, limiting the analysis to just the sentiment towards political actors can miss important factors of bias like ideology. 

However, despite the benefits of sentiment and the recent research into computational tools for sentiment classification, sentiment alone has limited utility in understanding more contextual attitudes and opinions, like stance. Stance detection entails the automated prediction of an author's viewpoint or stance towards a subject of interest, often referred to as the "target" \cite{alturayeif2023systematic}. Typically, a stance towards a subject is categorized as "Agree", "Disagree", or "Neutral". However, the labels representing stance can vary based on the specific target or context. Essentially, a stance mirrors an individual's perspective toward a specific topic or entity. Because stance inherently requires context in order to classify, stance detection remains a challenge for computational tools, especially those relying on keywords or supervised machine learning \cite{ng2022my, allaway2023zero}. Despite these challenges, there are a few, very recent works that show stance classification can be done without labeled data --- in an unsupervised or zero-shot setting --- in much the same way sentiment classification is currently done \cite{liyanage2023gpt, mets2023automated, zhang2022would, cruickshank2023use}. 

When considering methods to characterize cable news bias, social science methods can be useful but continue to have a human reliance that limits the potential scope and depends on subjective judgment. Supervised learning methods are also insufficient because we lack objective labels for the bias of transcripts to train on. Additionally, the use of just frame or sentiment analysis in computational techniques oftentimes requires selecting articles discussing a narrow range of topics. This limitation becomes more apparent when attempting to apply techniques to cable news, which can oftentimes have a broad range of speakers and topics within a single show. So, this study aims to integrate topic modeling with previously used sentiment analysis techniques to create a more dynamic form of media analysis. Doing so allows for cable news bias to be characterized across topics and for shifts over time to be captured.

\section{Methodology}
\label{sec:methodology}
In order to characterize the media bias of cable news transcripts, we devised a four-step procedure. The procedure begins with data cleaning, then transcript analysis, and then concludes with program comparison and clustering. Throughout these steps, the key dimensions of analysis were what is being said on cable news (identified through named entity recognition) and how those topics are portrayed (classified through stance analysis). The following figure, Figure \ref{fig:Methods_diagram}, summarizes the proposed method.

\begin{figure}[h]
\centering
\includegraphics[width = 0.45\textwidth]{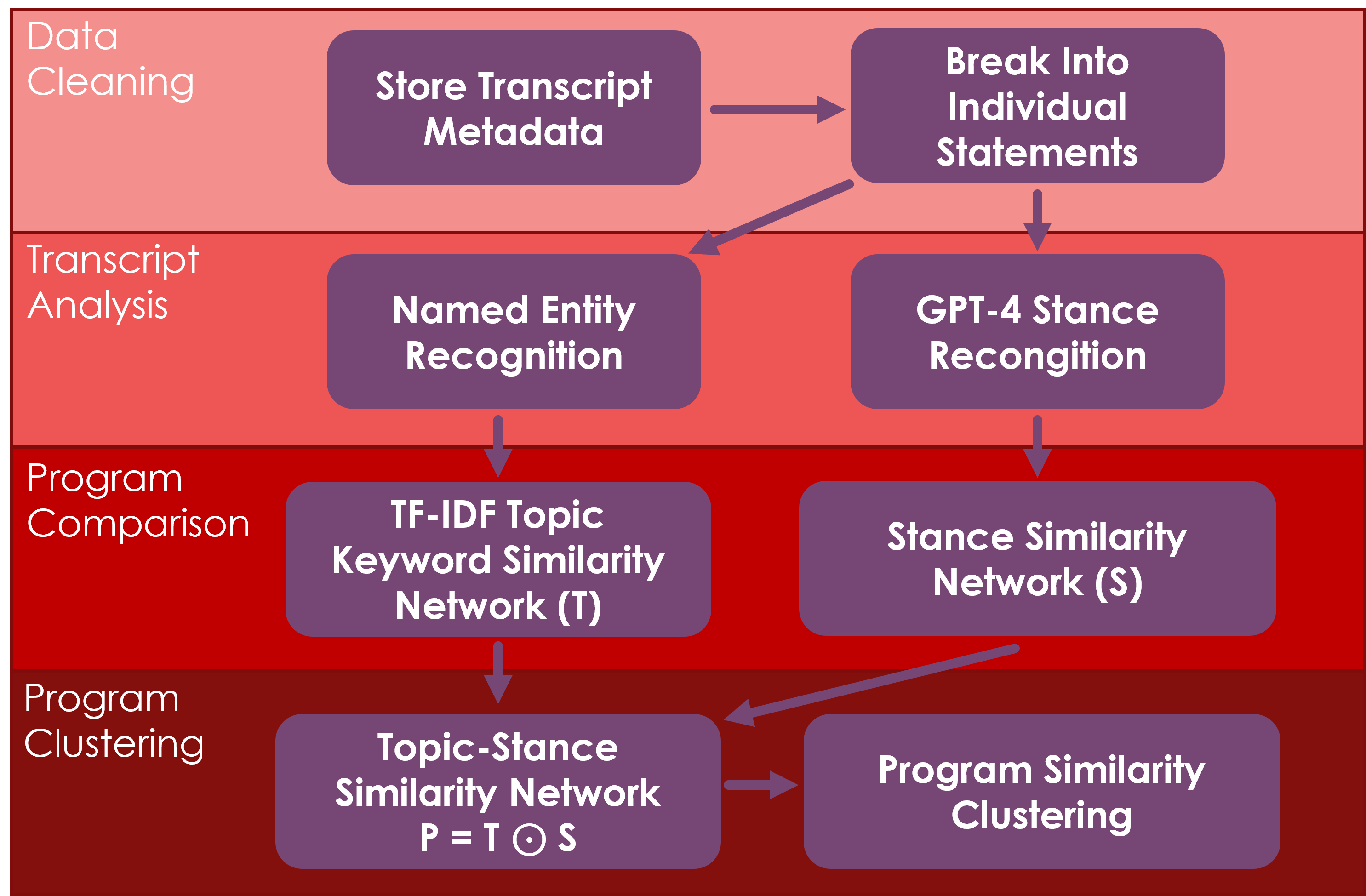}
\caption{Methods Diagram}
\label{fig:Methods_diagram}
\end{figure}

\subsection{Transcript Cleaning}
Using NexisUni, we obtained all transcripts from Fox News, MSNBC, and CNN from January 2020 to December 2020, a total of 14,000 transcripts. These networks were selected because they are thought of as the major networks in cable television and their transcripts had been used in previous cable news research \cite{benson2023campaigning}.  Using the information on the first page of each transcript, which was detailed in a consistent way throughout all transcripts, we retrieved metadata including the program name and the date it aired. Afterward, transcripts were parsed to create a collection of statements and associated speakers.  This allowed for a list of all statements made in a transcript to be collectively analyzed. Using the recorded date, the transcripts were divided by month so that the program analysis could be done at a monthly level and results could be compared across months. Since most news cycles occur between a weekly and monthly time frame, we choose to analyze at the monthly level but note the procedure could be done at any time scale. The month level was selected so that each program could have sufficient associated transcripts for analysis.

\begin{figure}[h]
\caption{Example Transcript First Page. The transcripts of the cable news shows come in a PDF format, which must be processed to extract the text.}
\centering
\includegraphics[width = 0.45\textwidth]{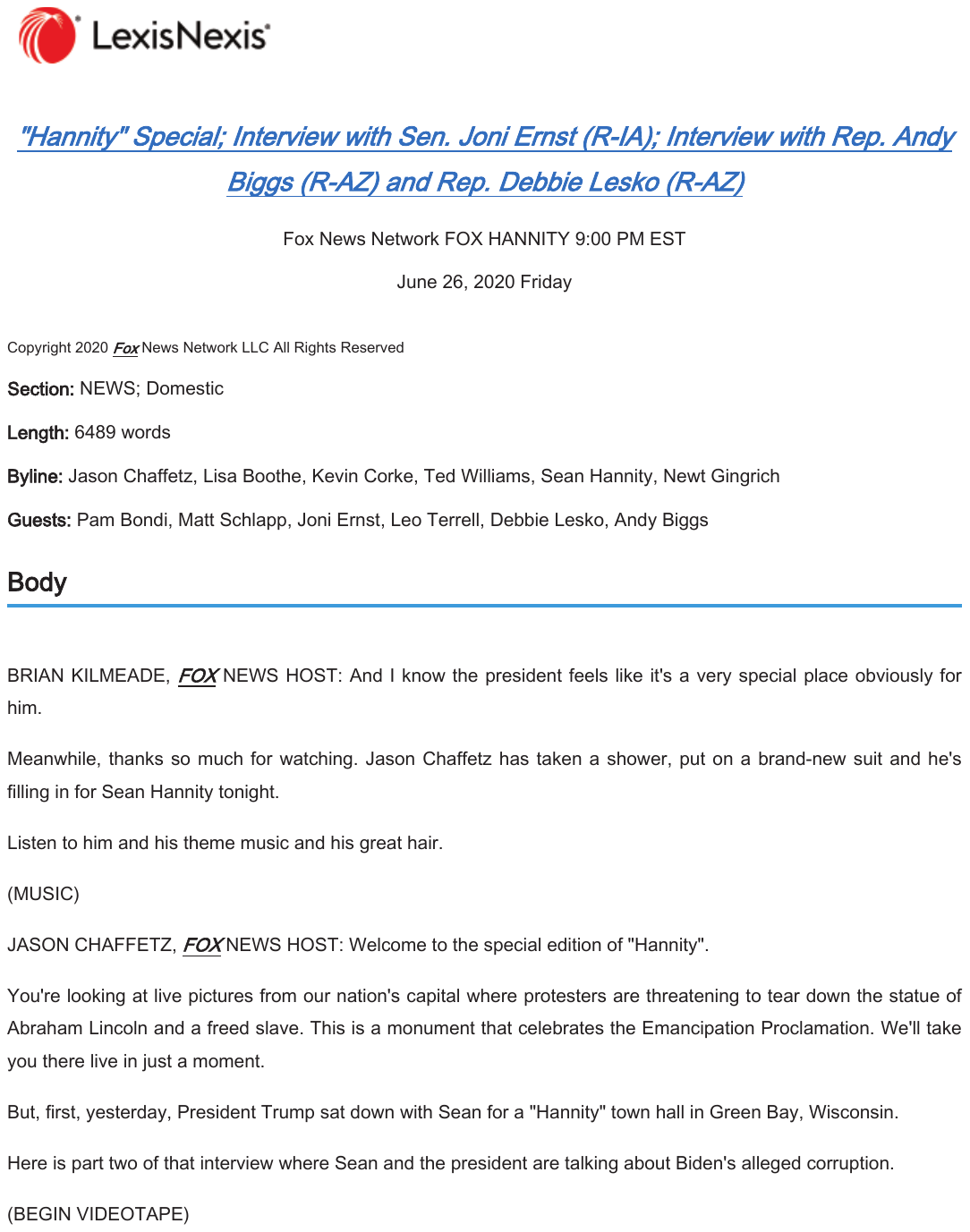}
\label{fig:transcript_example}
\end{figure}

\subsection{Named Entity Recognition}

The first part of transcript analysis was determining what is being said on cable news through named entity recognition. This is done through the EntityRecognizer method in the spaCy python package \cite{entityrecognizer}. The EntityRecongizer method identifies proper words or phrases within a text such as names, places, or organizations. We sort these by frequency within transcripts, cutting entities that are likely not news topics such as the time, quantities, cardinal directions, and percentages. All named entities that are in the top five entities in a transcript and are used in at least three sentences are marked as keywords for that transcript. Entities outside of these criteria are typically not main topics in a transcript. Additionally, entity limitations were required to limit the API requests and time required for the stance analysis.

Before deciding on named entity recognition as the means of discerning what is being talked about, we also investigated different approaches for topic identification. One was using topic modeling to split transcripts into multiple topics and take the most used non-common words within each topic as keywords by both LDA \cite{blei2003latent} and BERTopic \cite{grootendorst2022bertopic}. The other approach was a broad approach that selected the most frequently used nouns within each transcript as keywords. Each of these approaches returned similar results but they were less effective at differentiating programs and creating consistent clusters than by using the top entities from the transcripts.

\subsection{Stance Analysis}

After determining the keywords in a cable news program we turned to determining the expressed stance towards those words. We opted to use GPT-4 \cite{openai2023gpt4} due to it being state-of-the-art at the time of this research to accomplish this, but note that other LLMs could also be used for this analysis \cite{cruickshank2023use, liyanage2023gpt}. For each keyword identified in a transcript, we determine the transcript's stance towards that keyword to be the average of the individual sentence stances towards the keyword across all of the sentences it is the main subject of. Sentence level analysis was used because news programs often switch between several topics over the course of any given person's statement (e.g. a newscaster presenting an opinion monologue), so analyzing at the sentence level gives a better understanding of the sentiment toward any one topic. We use one prompt to determine both whether a keyword is the main subject of a sentence and what the sentence stance is towards the keyword if it is a main subject. For each sentence, GPT-4 was asked to "respond NO" if the keyword is not "the main subject in the sentence" and "return whether the statements are POSITIVE, NEUTRAL, or NEGATIVE towards" the subject if it is the main subject. After receiving the GPT-4 output, these results are quantitatively mapped so that POSITIVE equals a stance of one, NEUTRAL equals a stance of zero, and NEGATIVE equals a stance of a negative one. The average of these returned values is stored as the transcript's overall stance towards a keyword topic.

Prior to using stance, we calculated the sentiment for each sentence in a transcript using the VADER Composite Polarity score\cite{hutto_vader_2014}. VADER sentiment analysis uses the full context of a text, including punctuation and capitalization, to determine its sentiment \cite{hutto_vader_2014}. However, because VADER can only return how positive or negative text in general, not specifically towards a subject, it proved to be much less effective at differentiating programs than stance analysis. Evidence is shown in the results section.

\subsection{Program Comparison}

Completing the transcript analysis results in a data frame of keyword use, keyword sentiment, and the news program associated with the transcript. From there, we create two program-to-topic networks. The first is a topic frequency network ($B_{p,t}$), which portrays the number of times topic $t$ was an identified keyword out of program $p$'s transcripts. The second is a topic sentiment network ($C_{p,t}$), which contains the average stance of transcripts from program $p$ towards topic $t$. 

To create a program-to-program topic similarity network, we transform $B_{p,t}$. We first weigh for word relevancy by applying Term Frequency - Inverse Document Frequency (TF-IDF) \cite{sparck_jones_statistical_1972}. We then compare programs to each other using cosine similarity. The resulting network, $T_{p,p}$, portrays the similarities between programs based on their usage of keywords.

Next, we create a program-to-program stance similarity network by analyzing the similarities in sentiment towards shared keywords between programs. Because our stance scores range from -1 to 1, the largest potential distance in sentiment is 2. Given this, we calculate the stance similarity towards topic $t$ between programs $p_1$ and $p_2$ with the following equation:

$$s = \frac{2 - abs(C_{p_1,t} - C_{p_2,t})}{2} $$

$S_{p_1,p_2}$ is determined by taking the average stance similarity between program $p_1$ and program $p_2$ for all keywords they use at least once. So, if program $p_1$ and program $p_2$ share five keywords, $S_{p_1,p_2}$ will reflect the average of their similarity scores $s$ towards each of those five. By doing this for every program combination, we create $S_{p,p}$, which captures how similar programs are to each other based on their stance towards keywords.

\subsection{Program Clustering}

In order to combine topic and stance similarity between programs, we conduct elementwise multiplication between $T_{p,p}$ and $S_{p,p}$:

$$P_{p,p} = T_{p,p} \odot S_{p,p}$$

The created matrix $P_{p,p}$ represents the similarity between cable news programs across both topic and stance. We then want to group programs that are near each other by creating clusters. Since the nearness of the programs in $P_{p,p}$ is a result of the elements of bias, we can characterize the resulting clusters as bias clusters, wherein those programs clustered together share the same bias. We choose to use spectral clustering \cite{ng_spectral_2001}, which fits our data well because $P_{p,p}$ is based on the similarity between programs. We fit the clustering algorithm to the latent space representation of $P_{p,p}$ created with spectral embedding \cite{luo2003spectral}. For every month, a new network $P_{p,p}$ is created and each program is assigned to a cluster. Afterward, these clustering assignments are analyzed for two main factors: their consistency over time and their relationship with the cable news networks the programs are on.

\section{Results}
\label{sec:results}

\subsection{Monthly Program Aggregation}

\begin{figure}[h]
\caption{Section of April 2020 Topic Frequency Network}
\centering
\includegraphics[width = 0.45\textwidth]{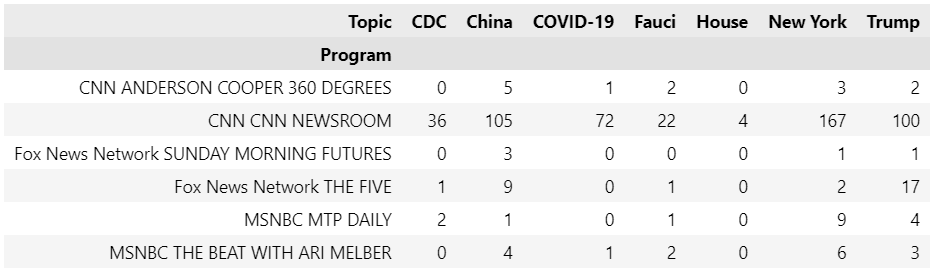}
\label{fig:april_example}
\end{figure}

This section examines the results from one month of analysis. We selected April of 2020 to demonstrate monthly output from our method. Figure \ref{fig:april_example} shows a section of $B_{p,t}$ for March, the frequency of certain topics on a selection of programs. It is apparent that some topics are discussed much more frequently (see "Trump") than others (see "House"), and also that some programs discuss certain topics much more than others. Note that programs, like CNN Newsroom, with more transcripts and longer airtimes typically mention topics more frequently over a month. 

\begin{figure}[h]
\caption{Section of April 2020 Topic Sentiment Network}
\centering
\includegraphics[width = 0.45\textwidth]{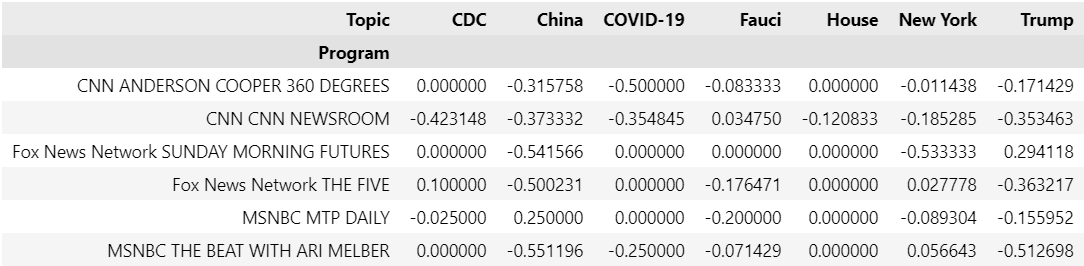}
\label{fig:stance}
\end{figure}

Next, we examined the average stance toward these keywords. A portion of $C_{p,t}$ is shown in Figure \ref{fig:stance}. Note that a zero value represents program-topic combinations where a topic never appeared as one of the program's transcript keywords. There is some variety in the average stance towards keywords, although largely in line with what we would expect. For example, The Beat with Ari Melber on MSNBC had the lowest average stance towards "Trump", which meets the typical assumption that MSNBC has portrayed Donald Trump unfavorably in its coverage.

\begin{figure}[h]
\caption{Latent Space Representation of Programs ($P_{p,p}$) in April 2020}
\centering
\includegraphics[width = 0.45\textwidth]{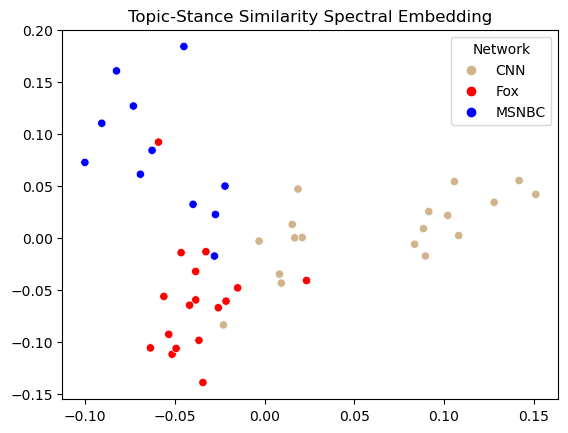}
\label{fig:spectral}
\end{figure}

Finally, we use spectral embedding to create a latent space representation of $P_{p,p}$.\cite{luo2003spectral} Because we use spectral clustering as our clustering algorithm, spectral embedding provides a visualization of how the clusters are selected in our model. The latent space representation, in Figure \ref{fig:spectral}, clearly depicts clusters that align with cable news networks, as almost all CNN, Fox, and MSNBC programs appear grouped together. Additionally, there are some outliers that are near programs from other networks, and CNN programs seem to be divided into two groups. 

Looking at the monthly level we also observe that the program clusters generally follow the networks. Table \ref{tab:adjusted_rand_index}, which depicts the Adjusted Rand Index between clustering assignments and network, shows that this is the case for most months. Other than in January, each month has an Adjusted Rand Index greater than 0.35, indicating an association between a program's network and its clustering assignment in our model.

\begin{table}[h]  
\centering  
\caption{Adjusted Rand Index of Program Clusters to Networks by Month}  
\label{tab:adjusted_rand_index}  
\begin{tabular}{lc}  
\toprule  
Month & Adj Rand Index \\   
\midrule  
January & 0.056  \\   
February & 0.369  \\   
March & 0.397  \\   
April & 0.411  \\   
May & 0.491  \\   
June & 0.434  \\   
July & 0.390  \\   
August & 0.563  \\   
September & 0.538  \\   
October & 0.710  \\   
November & 0.534  \\   
December & 0.342  \\   
\bottomrule  
\end{tabular}  
\end{table}  

\subsection{Full Year Results}

\begin{figure*}[!ht]
  \caption{Sankey of Assigned Clusters Across 2020. \textcolor{blue}{Blue} Indicates MSNBC Programs, \textcolor{tan}{Tan} CNN Programs, and \textcolor{red}{Red} Fox Programs.}
  \hspace*{-0.15in}
  \includegraphics[width=.95\textwidth]{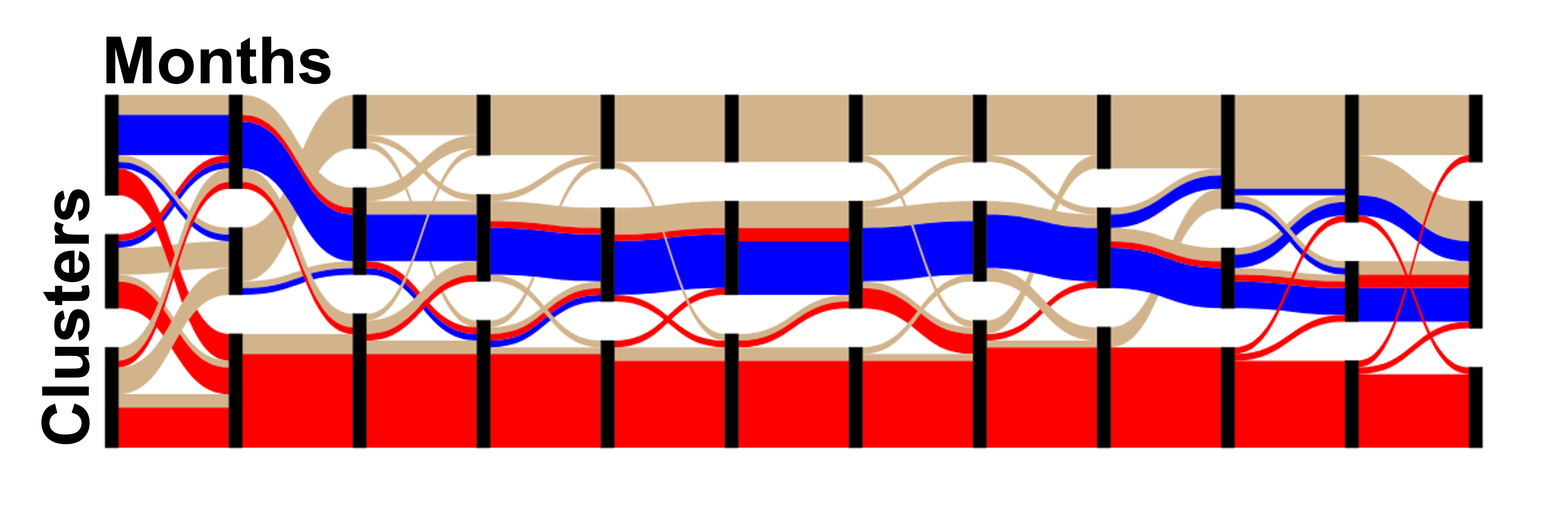}
  \label{fig:sankey}
\end{figure*}

Our next results examine clusters across the year 2020. Figure \ref{fig:sankey} contains a Sankey diagram of all clustering assignments of programs across months. In this diagram, each individual line on the verticle axis represents a cluster, while moving from right to left represents going from January to December 2020. Two things are apparent. First, the clusters are consistent over time. Although there are a few periods of major movement between months (especially in the first two months), most months see very few programs change clusters. Between March and December, there are rarely more than a couple of transcripts changing clusters at a time. The programs that do frequently move tend to be those typically less thought of as partisan, such as "Fox News Sunday".

Like the single months' results, we also see over time that the clustering assignments are strongly tied to networks. With a few exceptions, almost all of the red Fox programs are in the same cluster throughout the year. Similarly, the blue MSNBC programs are largely assigned to the same cluster over time. The tan CNN programs are also mostly together, although several cross over to the MSNBC-dominated middle cluster at different points of the year. 

The Sankey diagram shows programs grouped by cluster and network, but we also want to determine how individual programs compare to each other in their clustering assignments. Using Principle Component Analysis (PCA), we reduce the clustering assignments from each month into two dimensions.

\begin{figure}[h]
  \caption{PCA Representation of 2020 Clustering Assignments}
  \centering
  \includegraphics[width=.4\textwidth]{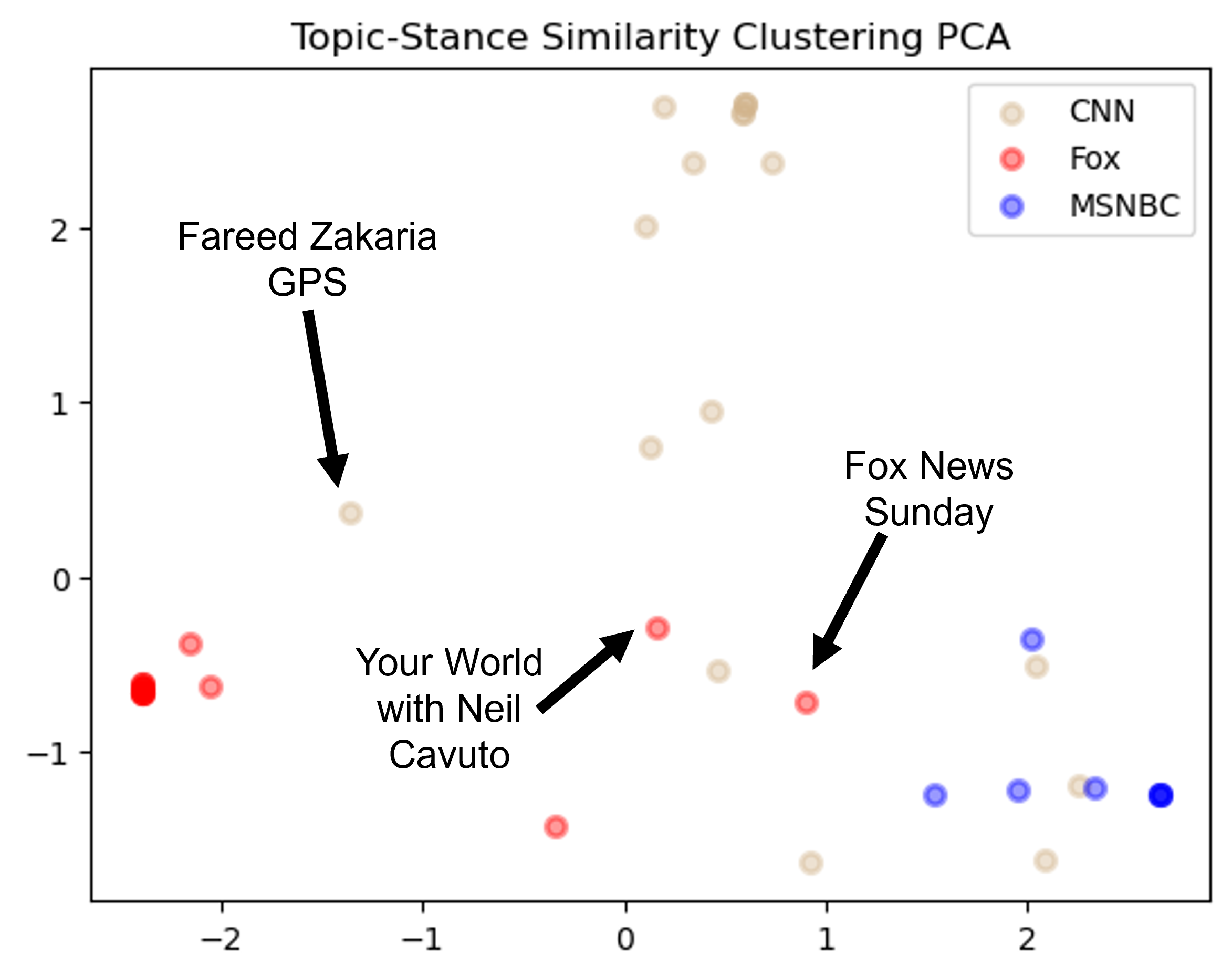}
  \label{fig:pca}
\end{figure}

Like the Sankey diagram, the PCA plot, Figure \ref{fig:pca}, reveals that programs are largely grouped together by network. Most of the Fox programs are tightly grouped together, with a few exceptions like Fox News Sunday placed closer to CNN and MSNBC programs. Opposite the Fox programs on the plot, MSNBC programs are also closely placed. Interestingly, CNN programs are more spread out, with some programs intermingled with the group of MSNBC programs while others are placed away from both the Fox and MSNBC groups. However, having many CNN and MSNBC programs placed closely together aligns with the conventional belief that CNN and MSNBC are similar in ideology and lean in the opposite direction of Fox.

\subsection{Effectiveness Sentiment versus Stance in Bias Determination}

As mentioned in the results section, using sentiment similarity to measure how programs talk about topics failed to produce meaningful variation between programs. On the other hand, stance similarity produced a much greater distinction between programs. One way to measure this is through the similarity matrix standard deviation. Given that some cable news programs are much closer to each other than others in their views on key topics and very different from others, we would expect there to be clusters of high similarity and large regions of low similarity. On the other hand, if a similarity matrix has values that are small in magnitude and close together, it would indicate that the measure of similarity is unable to distinguish between programs. Thus, having a higher similarity matrix standard deviation indicates that a model is better at capturing real-world conditions. Sentiment similarity resulted in a standard deviation (0.05 on average) that was only a third as large as the topic similarity standard deviation (0.15 on average), indicating that it captured much less variation between programs. Meanwhile, using stance resulted in a standard deviation (0.16 on average) roughly equal to the topic similarity standard deviation. 

\begin{figure}
  \begin{subfigure}{\linewidth}
  \includegraphics[width=.5\linewidth]{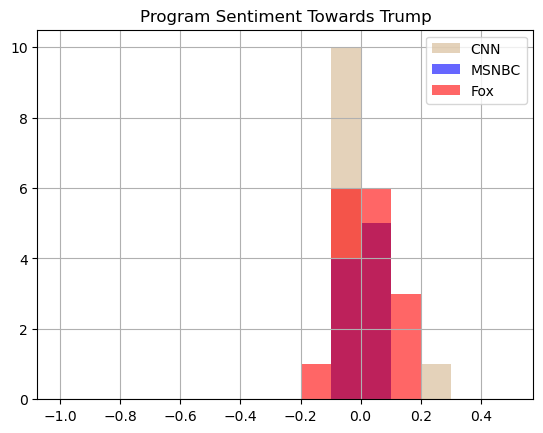}\hfill
  \includegraphics[width=.5\linewidth]{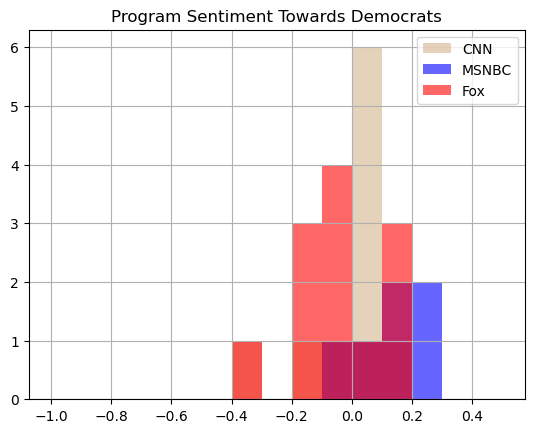}
  \caption{Sentiment Analysis Results}
  \end{subfigure}\par\medskip
  \begin{subfigure}{\linewidth}
  \includegraphics[width=.5\linewidth]{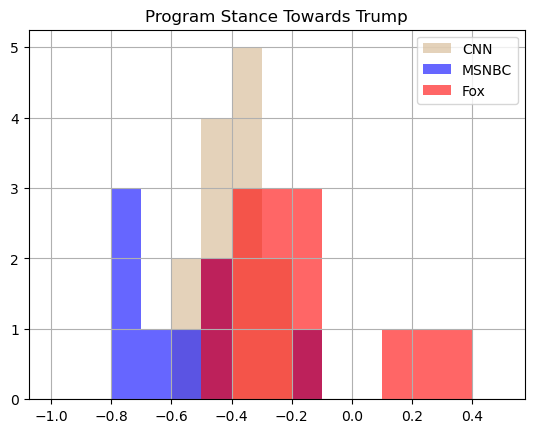}\hfill
  \includegraphics[width=.5\linewidth]{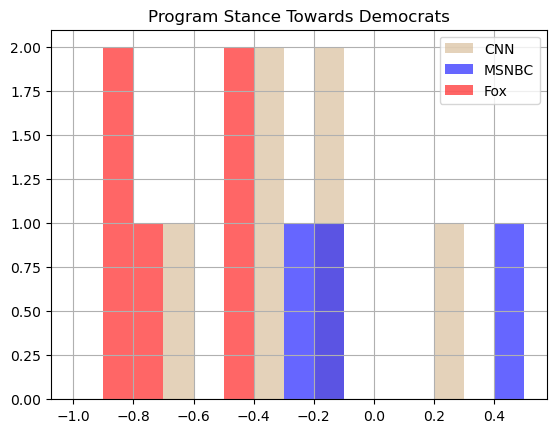}
  \caption{Stance Analysis Results}
  \end{subfigure}\par\medskip
  \caption{Program Sentiment and Stance Distribution for "Trump" and "Democrats" in April 2020}
  \label{fig:stance_vs_sentiment}
\end{figure}

Another way to compare performance between the methods is to look at the variance of programs' sentiment and stance towards specific topics. \ref{fig:stance_vs_sentiment} shows the distribution of program sentiment and stance towards "Trump" and "Democrats" in April 2020. Because these are highly partisan terms, we would expect the networks to differ in their coverage of them. We can see that this is not the case for sentiment, as programs from all three networks are tightly grouped together around the same average sentiment values. Comparatively, stance does a much better job of both providing variance between programs and sorting by network. The distribution for both terms covers a much wider numerical range than the sentiment results and there appears to be some sorting by the network, with CNN and MSNBC transcripts being generally more negative than Fox towards "Trump" and more positive than Fox towards "Democrats". While the variance and network sorting produced through stance might not completely match expectations, they are much more meaningful in separating out bias-driven opinions toward entities than the sentiment results. 

These results suggest that local sentiment does not properly identify differences in how topics are discussed on cable news. Sentiment's weakness might be because different programs discussing the same topic are directing similar sentiments in opposing directions. For example, the phrases "the investigation into President Trump is fraudulent" and  "the investigation reveals President Trump's failures" have similar sentiments but frame the same figure very differently. Consequently, our results also reveal the relative strength of stance analysis and suggest that it is the better approach for analyzing political text.

\section{Discussion}
\label{sec:Discussion}

This research yielded several significant results. Firstly, our findings suggest that it is feasible to computationally extract media bias characterizations with minimal human bias assessments. The empirical examination of news programs from three cable news networks, each with different perceived biases, generally confirmed broadly held assumptions about these programs. The biases exhibited by these programs remained consistent over time and strongly aligned with their respective networks. However, the key contribution of our work is not merely the affirmation of common media pundit observations but the fact that we reached this conclusion through an automated machine-learning model. This model required minimal human input and did not necessitate a priori bias specifications. While our findings still necessitate human interpretation of the resulting clusters, they are largely free from subjective human assessments and do not require media expertise for operation. The objective nature of this model demonstrates its potential to replace current media evaluation methods, which still heavily rely on human interpretation and assumptions. It also provides a tool for exploring realms where biases are not yet known, such as social media user biases. For political science researchers, this paper offers a novel framework for analyzing media bias and a confirmation of many implicit assumptions about biases in cable news.

Another significant finding of this research is the application of stance analysis methods to cable news transcript text. Our results indicated that stance analysis, employing task-based prompting with GPT-4, was far more effective than sentiment analysis in characterizing programs' attitudes towards topics. Our stance analysis method generated more variance in programs' views towards topics and produced variance that better aligned with real-world expectations of programs' viewpoints. This method not only offers an improvement for understanding cable news transcripts, but it also demonstrates a more advanced way to identify writing style bias in media generally.

However, the model presented in this paper does have certain limitations. It identifies key topics within transcripts as the most frequently used named entities, which may overlook certain types of topics, such as policy discussions. Furthermore, while stance detection was a significant improvement over sentiment analysis, the stance labeling of sentences within transcripts was not always accurate. The extent of these inaccuracies remains undetermined due to the absence of a labeled dataset for the cable news transcripts we utilized. Nevertheless, it appears that certain specific subsets, like short sentences lacking key context, resulted in a higher amount of incorrect labeling. The improvement of stance detection, in both a zero-shot setting like this and on spoken-word text that comprises news transcripts, remains a crucial area for future research.

\section{Conclusion}
\label{sec:conclusion}

The primary objective of this paper was to develop a model capable of characterizing the biases of cable news programs given a large volume of text data in the form of transcripts. Our focus was on analyzing gatekeeping bias, which pertains to the topics discussed on cable news programs, and writing style bias, which refers to the language used to discuss these topics. To achieve this, we dissected individual transcripts using Named Entity Recognition and Few-Shot Stance Detection, before employing Spectral Embedding and Clustering to group similar programs. Our results largely conformed to common expectations about cable news: cable news programs exhibit consistent biases that generally align with other programs on their network.

Future research could leverage different models or prompting techniques to find improved ways to identify the stance of cable news text towards topics. Beyond our model, there are also other dimensions of bias that merit investigation. For instance, integrating our model with work done on visual bias in television could potentially enhance its ability to characterize bias in cable news \cite{coleman_network_2006}. Future work could also aim to examine a broader time range. This could reveal changes in cable news programs over time and potentially identify years that do not exhibit the consistent network-driven clusters we identified in the 2020 data.

\bibliographystyle{unsrt}
\bibliography{MA491}
\EOD

\end{document}